\documentclass{article}
\usepackage{spconf,amsmath,graphicx}

\usepackage{cite}
\usepackage{amsmath,amssymb,amsfonts}
\usepackage{algorithmic}
\usepackage{graphicx}
\usepackage{textcomp}
\usepackage{url}
\usepackage{subfig}
\def\BibTeX{{\rm B\kern-.05em{\sc i\kern-.025em b}\kern-.08em
    T\kern-.1667em\lower.7ex\hbox{E}\kern-.125emX}}

\begin{document}

\title{A New Cervical Cytology Dataset for Nucleus Detection and Image Classification (Cervix93) and Methods for Cervical Nucleus Detection\\
}

\twoauthors
  {Hady Ahmady Phoulady}
	{University of Southern Maine\\
	Department of Computer Science\\
	Portland, ME}
  {Peter R. Mouton}
	{University of South Florida\\
	Department of Computer Science and Engineering;\\ SRC Biosciences\\
	Tampa, FL}


\maketitle

\begin{abstract}
Analyzing Pap cytology slides is an important tasks in detecting and grading precancerous and cancerous cervical cancer stages. Processing cytology images usually involve segmenting nuclei and overlapping cells. We introduce a cervical cytology dataset that can be used to evaluate nucleus detection, as well as image classification methods in the cytology image processing area. This dataset contains 93 real image stacks with their grade labels and manually annotated nuclei within images. We also present two methods: a baseline method based on a previously proposed approach, and a deep learning method, and compare their results with other state-of-the-art methods. Both the baseline method and the deep learning method outperform other state-of-the-art methods by significant margins. Along with the dataset, we publicly make the evaluation code and the baseline method available to download for further benchmarking.
\end{abstract}


\section{Introduction}
\begin{figure}[!t]
\centering
\subfloat[Negative]{\includegraphics[width=1.1in]{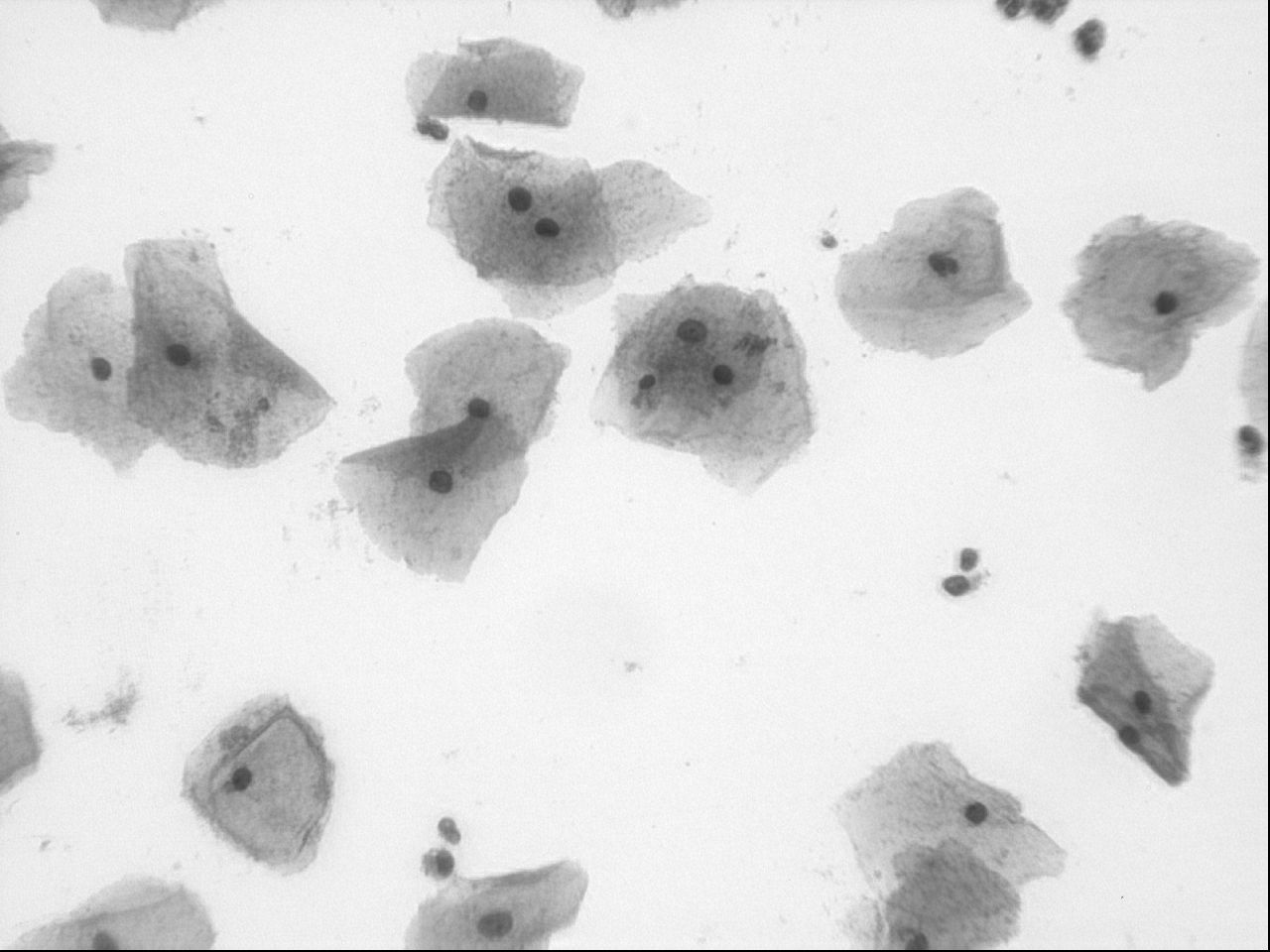}%
\label{n}}
\hfil
\subfloat[LSIL]{\includegraphics[width=1.1in]{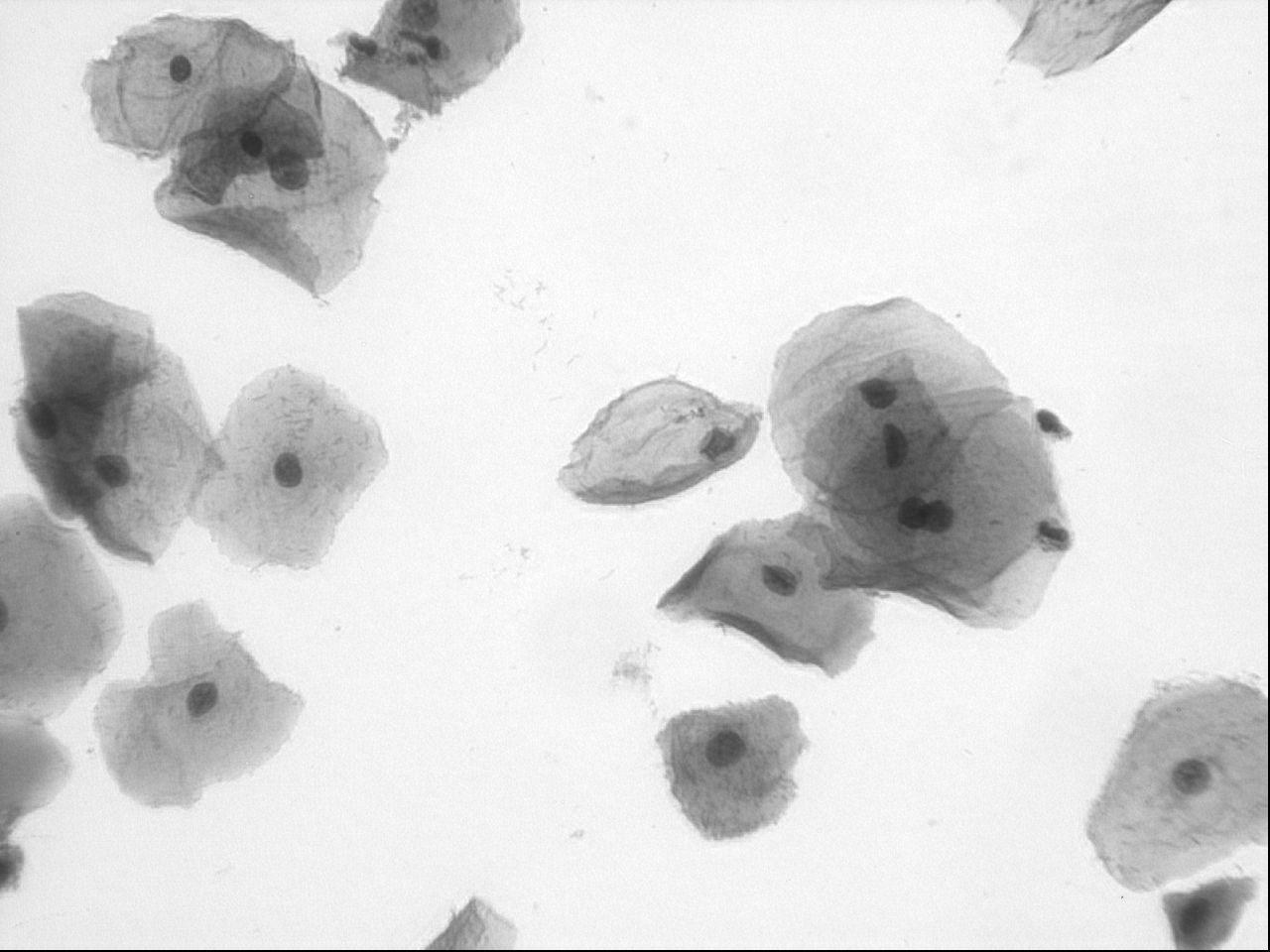}%
\label{lsil}}
\hfil
\subfloat[HSIL]{\includegraphics[width=1.1in]{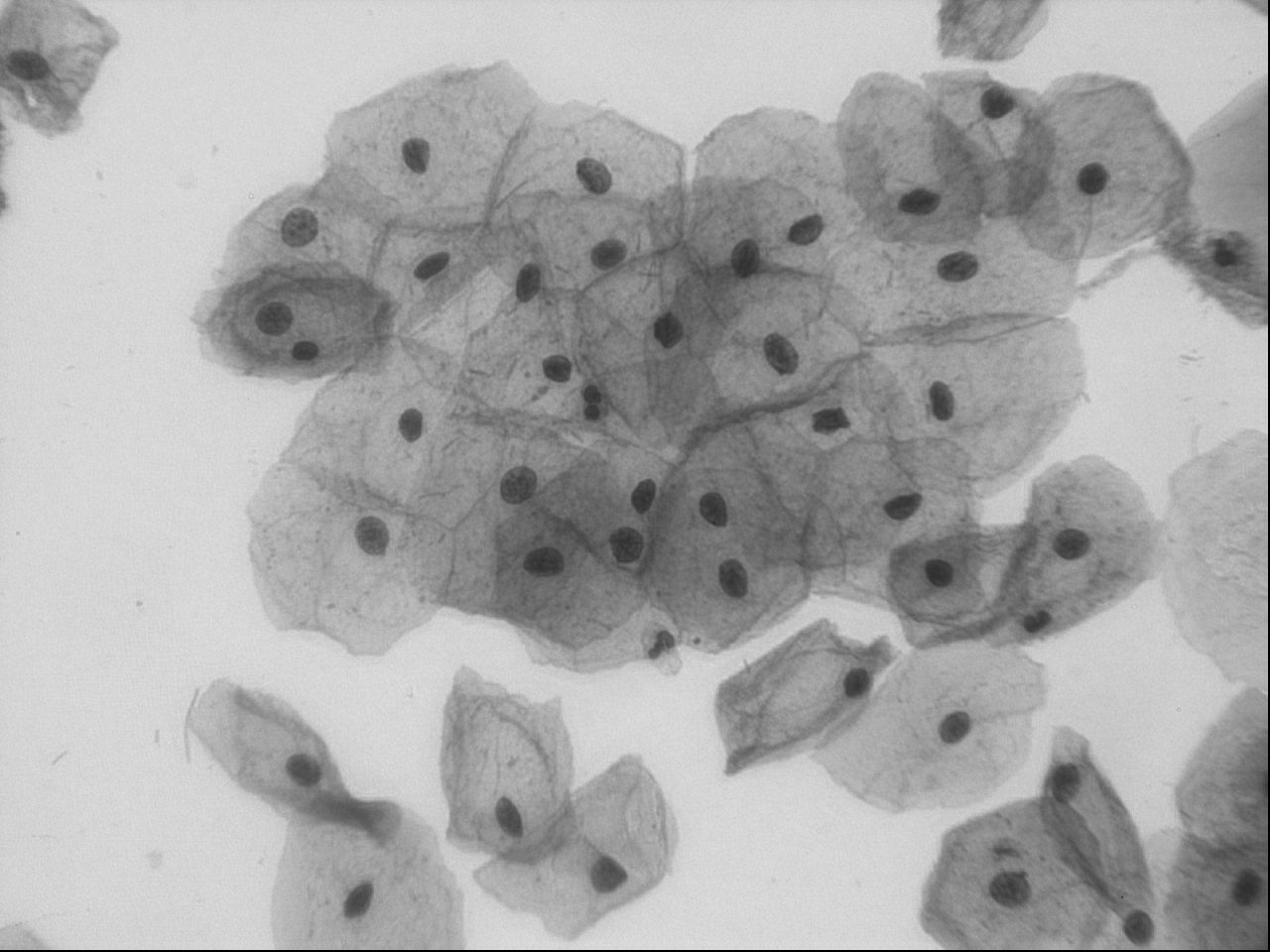}%
\label{hsil}}
\caption{Frames from each of the grade categories.}
\label{fig:dataset}
\end{figure}

Cervical cancer incidence and mortality rates have decreased by more than 50\% over the past three decades with most of the reduction attributed to screening with the Papanicolaou (Pap) test to detect cervical cancer and precancerous lesions \cite{14}. According to the Bethesda System (TBS) \cite{bethesda}, a Pap test is negative if there are no epithelial cell abnormalities. In the past three year, there have been an increase in the number of studies on automating the analysis of cervical cytology slides \cite{tareef2017optimizing,kumar2016unsupervised,mine4,islam2015multi,gautam2017unsupervised,saha2017circular,zhao2017novel,ISBI2014,ISBI2015}. Some of these methods aim to segment nuclei \cite{gautam2017unsupervised,saha2017circular} while the others segment overlapping cervical cells \cite{tareef2017optimizing,kumar2016unsupervised,mine4,islam2015multi,zhao2017novel,ISBI2014,ISBI2015}. In either case, all methods include a step to segment nuclei, which are mostly used as seed points for subsequent cell segmentation. The accuracy of nucleus segmentation and detection is therefore the most important tasks in automating the analysis and segmentation of cytology images. As a result, publicly available datasets are useful evaluating and comparing the performances of nuclear segmentation and detection methods. After two overlapping cell segmentation challenges in ISBI 2014 and 2015, two new datasets were made available for evaluating the accuracy of nucleus detection and nucleus/cell segmentation methods \cite{ISBI2014,ISBI2015}. However, the limited number of annotated nuclei and low variation of images in these datasets renders them difficult for evaluation of nucleus detection accuracy. In this work, we present a dataset that includes 93 real Extended Depth of Field (EDF) images along with their image stacks from slides with three different Pap test grades: Negative; Low-grade Squamous Intraepithelial Lesions (LSIL); or High-grade Squamous Intraepithelial Lesions (HSIL). A total of 2705 nuclei are marked (annotated) in the images and together with a grade label for each image, allowing the dataset to be used for evaluating the performance and accuracy of nucleus segmentation and detection. Importantly, the dataset can be used for cytology classification of each into different grade categories, which is one of the primary goals for the methods developed for processing images of cervical cytology. For assessing accuracy and comparison, we provide the evaluation code and a baseline method that currently outperforms two other state-of-the-art methods on the dataset \cite{ISBI2014,ISBI2015}. In this paper, we also present a nucleus detection method based on the Convolutional Neural Network (CNN). In the future, we intend to make the cytoplasm and nucleus boundary annotation available. Fig. \ref{fig:dataset} shows sample images from each of the grade categories.

\section{Dataset}
As can be seen in Section \ref{sec:exp}, many of the state-of-the-art methods fail to perform reasonably well on larger datasets of real images that show relatively high levels of variability. In large part this is due to the lack of a publicly available comprehensive dataset of cervical cytology images with annotations. To this end, a more challenging dataset of real cervical cytology images is provided. During the dataset preparation, the goal was to build a dataset that is highly representative of routine cervical cytology images in order to recapitulate the features of such datasets in the field.

Archived ThinPrep Pap-stained slides provided by the Moffitt Cancer Center (Tampa, FL) were used for these studies. Images from these slides were acquired in a systematic-random manner using an integrated hardware-software microscope system (Stereologer, SRC Biosciences, Tampa, FL). Each of the slides was examined microscopically microscope by the same cytotechnologist and graded based on The Bethesda System (TBS) \cite{bethesda}. The slides used in this work were either graded as Negative; Low-grade Squamous Intraepithelial Lesions (LSIL); or High-grade Squamous Intraepithelial Lesions (HSIL). The automatic XYZ stepping motor and Stereologer software were used to manually determine the top and bottom of each slide at a manually selected focal plane. All images were of size 1280x960 pixels. The dataset includes four main parts as explained below.

\begin{enumerate}
	\item Image stacks: 93 stacks of images at 40x magnification are included in the dataset. Each of the stacks have 10-20 images acquired at equally spaced field of views from top to the bottom of the slide (named frame000 -- frame092). 
	\item Extended Depth of Field (EDF) images: An EDF image was created from each stack by the method proposed in \cite{edf} and included in EDF folder.
	\item Frame labels and training/test partition: each of the frames are accompanied with the label N, L or H (for Negative, LSIL, or HSIL, respectively). Moreover, based on the slides that these frames were taken from and their visual difficulty level the whole set of frames were partitioned to training and test sets (roughly 25\% of frames from each label were put in the test set). Frame labels and training/test partitioning is included in \texttt{label.csv}, with 0 and 1 indicating training and test set, respectively.
	\item Manually marked points coordinates: each of the frames was examined and a point inside each cervical nucleus marked manually. A cervical nucleus touching the boundary or a cervical nucleus partially inside a frame was marked only if more than half of it was located in the frame and if the center of it was roughly about 10 pixels far from the boundary.
\end{enumerate}

\subsection{Summary}
As mentioned above the dataset includes 93 frames. Table \ref{tab:grades} shows the number of frames in each of the grades N, LSIL or HSIL. A total of 2705 nuclei were manually marked in all frames across all grade categories (Table \ref{tab:points}). The dataset is available to the public\footnote{The dataset and the codes can be downloaded at the following link: \url{https://github.com/parham-ap/cytology_dataset}} 
along with an evaluation code and a baseline segmentation method for nucleus detection as described in the next Section.

\begin{table}[!t]
\caption{Number of Frames within Each Grade Category}
\label{tab:grades}
\centering\small
\begin{tabular}{|c|c|c|c||c|}
	\hline
	& Negative & LSIL & HSIL & Total\\
	\hline
	Training & 12 & 34 & 23 & 69\\
	\hline
	Test & 4 & 12 & 8 & 24\\
	\hline\hline
	Total & 16 & 46 & 31 & 93\\
	\hline
\end{tabular}
\end{table}

\begin{table}[!t]
\caption{Number of Nuclei within Each Grade Category}
\label{tab:points}
\centering\small
\begin{tabular}{|c|c|c|c||c|}
	\hline
	& Negative & LSIL & HSIL & Total\\
	\hline
	Training & 179 & 1125 & 679 & 1983\\
	\hline
	Test & 59 & 411 & 252 & 722\\
	\hline\hline
	Total & 238 & 1536 & 931 & 2705\\
	\hline
\end{tabular}
\end{table}

\section{Evaluation Measures}
The evaluation code in MATLAB made available with the dataset includes: 1) a list of coordinates of detected points for nuclei; 2) a single binary segmentation mask; or, 3) a cell of binary masks (for each segmented nucleus). In case of a list of coordinates, a detection is counted True Positive (TP) if it is in vicinity of a point (less than 10 pixel far) in the ground truth. Every segmented region can result in at most one TP. A False Negative (FN) refers to a point in ground truth that does not fall within any segmented regions. If more than one point falls inside a segmented region all points but one will be counted as FN if the points are not covered by any other segmented region. Finally, a False Positive (FP) is any detected point that is more than 10 pixels from a manual point or any segmented region that does not include a manual point. The evaluation metrics used to report the results are
\[
\text{Precesion = }\frac{\text{TP}}{\text{TP + FP}},\text{ and }\text{Recal = }\frac{\text{TP}}{\text{TP + FN}}.
\]
The Standard Deviation (STD) of Precision and Recall measures within each separate image are also computed by the code.

\begin{figure}[!t]
\centering
\subfloat[Negative]{\includegraphics[width=1.1in]{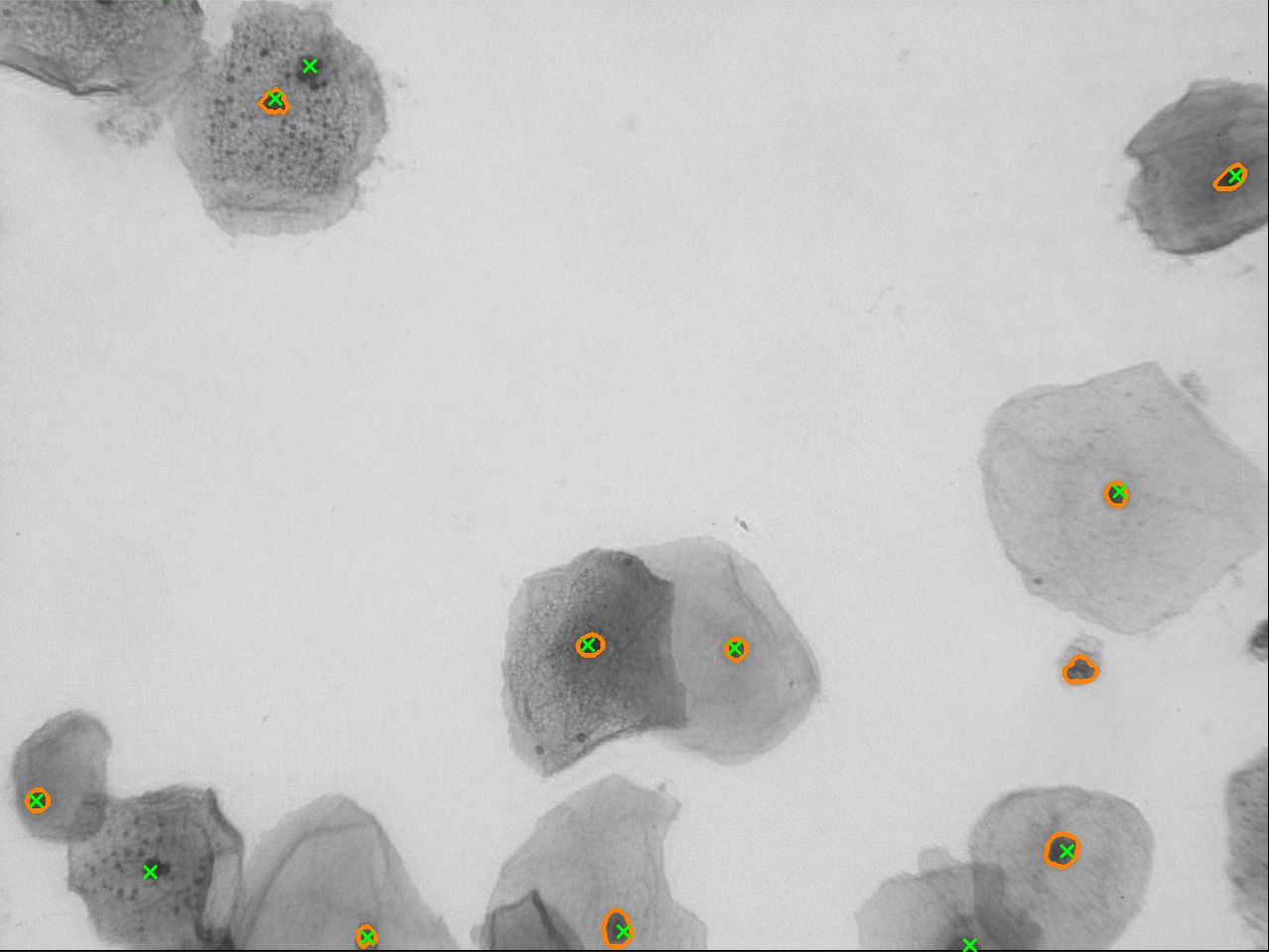}%
\label{rn}}
\hfil
\subfloat[LSIL]{\includegraphics[width=1.1in]{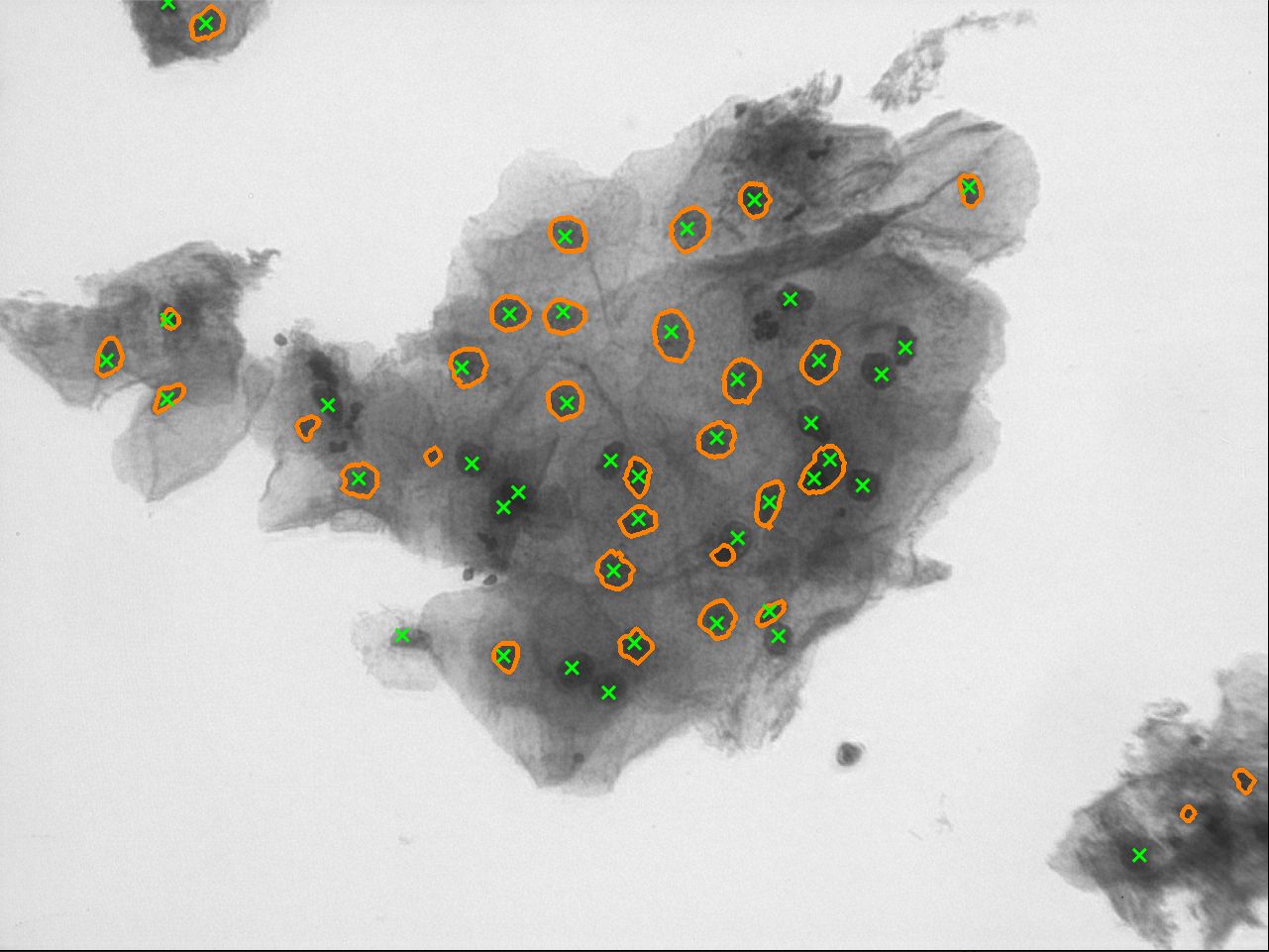}%
\label{rlsil}}
\hfil
\subfloat[HSIL]{\includegraphics[width=1.1in]{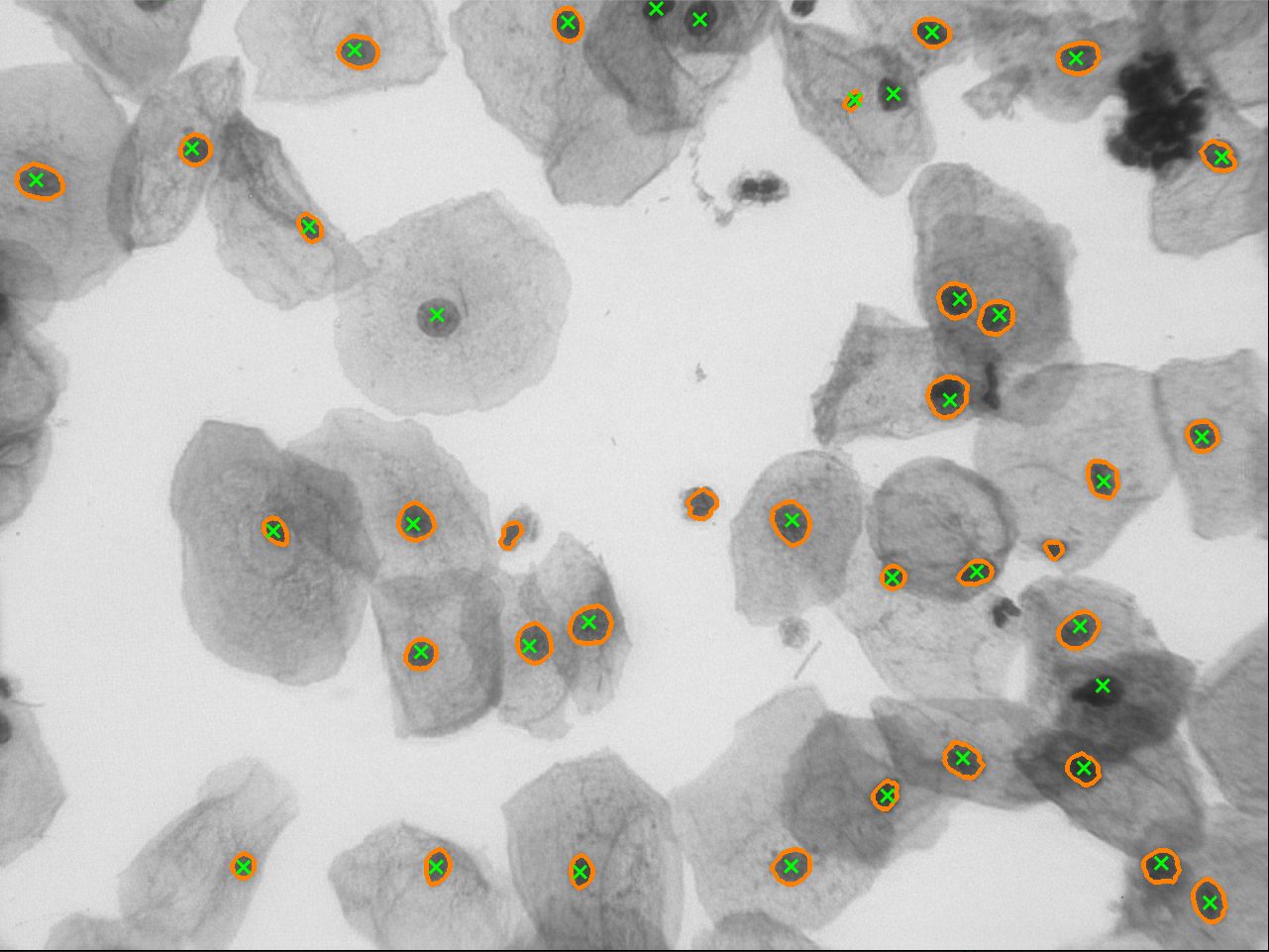}%
\label{rhsil}}
\caption{Segmentation results on three test frames. The green crosses are the manually marked points within each frame.}
\label{fig:results}
\end{figure}

\section{The Baseline Method}
We include a revised version of the nucleus segmentation algorithm proposed in \cite{phoulady2017framework} as a baseline method that segments cervical in the proposed dataset. The main differences are setting the parameters using the training set (instead of choosing them empirically as a large training dataset exists), removing the check for regions having a relatively low inside and outside average intensity difference (because it is rather an image specific characteristic), and adding the removal of regions overlapping with the boundary (as most of the nuclei touching the boundary are not annotated except they are mostly inside the frame). The segmentation result is then evaluated using the evaluation code and the results are discussed in Section \ref{sec:exp}.

The method first blurs the image with a 2-D adaptive noise-removal filter \cite{two_d_filter2} and iteratively binarizes the processed image with a set of increasing thresholds. The goal in the iterative thresholding was to find seed points inside nuclei and then grow the regions by increasing the threshold in the subsequent thresholdings. Regions that were too small or too concave (by measuring region's solidity) were removed in the iterative process. Moreover, an acceptable range of the average intensity within each nucleus was defined, and regions were allowed to merge if the new larger region was more \emph{solid} than all other separate regions. The the minimum and maximum accepted values for the average intensity along with the minimum size and the minimum acceptable solidity were trained and set using the training set in the dataset. Finally, the regions that were overlapping with the boundary were removed.

Unlike the method in \cite{phoulady2017framework}, the parameters were selected based on the method evaluation on the training set as explained below.

Different sets of parameters were explored using a grid search. The set of parameters resulting in the highest F measure,
\[
\frac{2 \cdot\text{Precision}\cdot\text{Recall}}{\text{Preceision + Recall}},
\]
on training set was selected to test the method's performance on the test set. The parameters that achieve the highest F measure on the training were used to test the method on the test set.


\section{The Deep Learning Method}
We trained and tested a Convolutional Neural Network (CNN) to detect cervical nuclei in the dataset. The designed CNN architecture has two convolutional layers (CL) and two fully connected layers (FCL). Each CL was followed by max pooling layers (MPL) and each FCL was followed by ReLu activation function (RL). Finally, the output layer (OL) with two neurons was followed by a soft max loss layer (SFL) to generate the probabilities of each patch being a nucleus or not.

Image patches of size 75x75 pixels were extracted from the grayscale frames in the training dataset. Patches were sampled uniformly at 15-pixel intervals and each of them was labeled as positive if it was within 15 pixels of a manually marked point; otherwise, it was labeled as negative. The network was trained with a constant learning rate of 0.001 in 100 epochs. Overall, 654,948 were extracted and used to train the network with only 1.71\% of patches labeled as positive.

The trained CNN was used to classify patches extracted from the frames in the test set. The patches were extracted at shorter interval compared to the training dataset, namely at 3 pixels apart. The hit map was generated, dilated, and thresholded at cut off point of 0.5. Small regions that did not represent the nuclei (regions smaller than 100 pixels) were removed.

\section{Experiments and Results}\label{sec:exp}
The parameters of the baseline method that achieve the highest F measure on the training set were 150, 10, 120 and 0.88 for minimum size, minimum average intensity, maximum average intensity and minimum solidity, respectively. Fig. \ref{fig:results} shows the visual results on one frame from each grade category. It has been observed (as it can also be seen in the sample segmentation results) that the method fails mostly on highly clustered areas with very low contrast. Some false negatives appear on the boundaries and false positives are mostly cause by artifacts and blood cells. The results of the revised method on the training and the test set is presented in Table \ref{tab:training_results}. The method fails to detect and segment about only 16\% of nuclei in the test set, and about 20\% of the segmented nuclei were False Positives.

\begin{table}[!h]
\caption{Results on the Training and Test Sets}
\label{tab:training_results}
\centering\small
\begin{tabular}{|c|c|c||c|}
	\hline
	& $\text{Precision}\pm\text{STD}$ & $\text{Recall}\pm\text{STD}$ & F\\
	\hline
	Training & $0.790\pm0.162$ & $0.792\pm0.135$ & $0.791$\\
	\hline
	Test & $0.803\pm0.137$ & $0.838\pm0.128$ & $0.820$\\
	\hline
\end{tabular}
\end{table}

\begin{table*}[!h]
\caption{Methods Performance on the Test Set (superior results are shown in bold font)}
\label{tab:results}
\centering\small
\begin{tabular}{|c|c|c||c|}
	\hline
	& $\text{Precision}\pm\text{STD}$ & $\text{Recall}\pm\text{STD}$ & F\\
	\hline
	Lu et al. \cite{ISBI2014} & $0.617\pm0.258$ & $0.295\pm0.274$ & $0.399$\\
	\hline
	Ushizima et al. \cite{ISBI2014} & $0.687\pm0.217$ & $0.446\pm0.230$ & $0.541$\\
	\hline
	Phoulady et al. \cite{phoulady2017framework} & $0.753\pm0.152$ & $0.716\pm0.186$ & $0.734$\\
	\hline
	Baseline Method & $0.803\pm0.137$ & $0.838\pm0.128$ & $0.820$\\
	\hline
	CNN & $\mathbf{0.861\pm0.095}$ & $\mathbf{0.895\pm0.086}$ & $\mathbf{0.878}$\\
	\hline
\end{tabular}
\end{table*}

We also present the results of the method in \cite{phoulady2017framework}, and two methods discussed in \cite{ISBI2014} proposed by Lu et al. and Ushizima et al. both in the ISBI 2014 challenge. As it can be seen from the results, the baseline method outperforms other methods by a significant margin. The original version of the method in \cite{phoulady2017framework} performs relatively well and achieves significantly better results that the two methods in \cite{ISBI2014}. This shows that the baseline method and its original version are more applicable to new datasets than the other two state-of-the-art methods.

Because of the adequate number of frames and nuclei in the dataset, we were able to propose, train and test a CNN that easily outperformed other methods and achieved significantly better results than the other methods (except for the baseline method with improvement of about 6\%). More sophisticated network architectures will potentially increase the accuracy of the method further. However, one of the main disadvantages of this method, in comparison to the baseline method is the longer time that it needed to train and test on new images. However, in terms of computation speed, testing the trained network on unseen data was relatively faster than the method proposed by Ushizima et al. and was significantly faster than the method proposed by Lu et al.


\section{Conclusion}
The introduced dataset contains 93 real cervical cytology images acquired from Normal, LSIL and HSIL graded ThinPrep cytology slides. 2705 manually marked points inside nuclei in the images along with image labels are provided as ground truth in the dataset. The dataset can be used to assess the performance of nucleus segmentation and detection, and image classification methods in this research area. 


This dataset is more comprehensive in terms of the number of manually marked nuclei, the number of real images and the number of slides with varying degree that was used to acquire the images. Two other state-of-the-art methods that achieve good results on ISBI 2014 and 2015 overlapping cervical cell segmentation challenge \cite{ISBI2014,ISBI2015} datasets perform poorly on this dataset. It shows that the new dataset is much more challenging for the task of nucleus detection and segmentation compare to other two datasets. The method in \cite{phoulady2017framework} achieves better results compared to the other two methods, but those results are still significantly lower than the results previously reported on the other datasets.

As it can be seen in Table \ref{tab:results}, the baseline method applied to the test set can achieve a precision and recall of 0.803 and 0.838, respectively. This is substantially lower than what the original method in \cite{phoulady2017framework} could achieve on the dataset introduced in ISBI 2014 and ISBI 2015 challenges. Specifically, on ISBI 2014 test set the method could achieve 0.874 and 0.930 precision and recall, respectively, due to the more difficult task of nucleus detection and segmentation in this dataset. Images in the test set of ISBI 2014 challenge were synthetic images and presented a limited variability. All images in this dataset are real images and the nuclei in different images present a much more variation compared to both ISBI 2014 and ISBI 2015 datasets. The best result was achieved by the proposed CNN. It improved the F measure obtained by the baseline method by more than 5\%. It only missed about 10\% of the nuclei. It has been observed that most of the missed nuclei were close to the boundary and/or were due to overlapping nuclei. These errors can be potentially reduced by a more advanced post-processing step.

In future work, our goal is to include the cytoplasm and nuclei segmentation ground truth in the dataset to enable researchers to contrast and compare the relative accuracy and performance of their segmentation methods. Furthermore, more sophisticated CNN architectures will be developed and investigated to improve the results. Finally, selecting frames from different grade categories allows the results of any segmentation method to extract first-order stereology parameters such as Mean Nuclear Volume. Thus, the methods developed with this dataset could be used for cervical cytology image classification based on abnormalities in nuclear size, a critical step toward automatic and objective quantification of the cervical cytology.

\section*{Acknowledgment}
We would like to thank Dr. Siegel at Moffitt Cancer Center for providing the Pap slides used to prepare this dataset.

\bibliographystyle{IEEEbib}
\bibliography{references}

\end{document}